\documentclass[11pt]{article}
 \RequirePackage{fix-cm}

\usepackage{amssymb, amsmath, amsfonts, mathtools}
\usepackage{epsfig}
\usepackage{graphicx}
\usepackage{subfigure}
\usepackage{epstopdf}
\usepackage{multicol}
\usepackage{array}
\usepackage{booktabs}

\usepackage{mathrsfs}
\usepackage{algorithm2e}
\usepackage {tikz}
\usetikzlibrary {positioning}
\usetikzlibrary{arrows}
\usetikzlibrary{decorations.pathreplacing}
\usepackage {xcolor}

\usepackage{tikz-qtree}
\usepackage{latexsym}

\definecolor {processblue}{cmyk}{0.96,0,0,0}

\newcolumntype{C}[1]{>{\centering\let\newline\\\arraybackslash\hspace{0pt}}m{#1}}
\newcolumntype{L}[1]{>{\raggedright\let\newline\\\arraybackslash\hspace{0pt}}m{#1}}

\graphicspath{{images/}}

\begin{document}

\title{Belief Base Revision for Further Improvement of Unified Answer Set Programming
}

\author{Kumar Sankar Ray\\
ECSU,Indian Statistical Institute, Kolkata
\and
Sandip Paul\\
ECSU,Indian Statistical Institute, Kolkata
\and
Diganta Saha\\
CSE Department\\
Jadavpur University}

\maketitle

\begin {abstract}

A belief base revision strategy is developed. The belief base is represented using Unified Answer Set Program which is capable of representing imprecise and uncertain information and perform nonomonotonic reasoning with them. The base revision operator is developed using Removed Set Revision strategy. The operator is characterized with respect to belief base revision postulates.

\end {abstract}

\section{Introduction}

In the domain of knowledge representation and reasoning, belief revision plays an important role. The objective of belief revision is to study the process of belief change; i.e., when a rational agent comes across some new information, which contradicts his or her present believes, he or she has to retract some of the beliefs in order to accommodate the new information consistently. The three main principles, on which the belief revision methodologies rely upon, are; 1. Success: The new information must be accepted in the revised set of beliefs; 2. Consistency: The set of beliefs obtained after revision must be consistent; 3. Minimal Change: In order to restore consistency, if some changes have to be incurred, then the change should be as little as possible.

The information of a rational agent can be represented by a deductively closed set of rules, i.e., a belief set, or by a set of rules that is not closed under consequence relation, i.e., a belief base. Belief set revision is characterized by means of AGM postulates \cite{gardenfors2003belief,pagnucco1996introduction} for propositional logic. Later AGM style belief revision has been extended for logic programs with answer set semantics \cite{delgrande2008belief,delgrande2013agm,delgrande2009merging,aravanis2017belief}. All of these approaches are based on distance-based belief revision operators constructed on SE models. However it is proved that these belief change operators suffer from some serious drawbacks \cite{slota2010semantic}. On the other hand, some syntactic approaches are available that deal with belief base merging or belief base revision. A belief base can be represented by a logic program as, in general, a logic program is a set of rules not deductively closed under the consequence relation. Belief base revision for ASP has been developed for quite sometime \cite{eiter2002properties}.

All the approaches mentioned above are based on classical two valued logic. However, in a real life scenario, belief bases may have inherent uncertainty and vagueness due to the incomplete and imprecise nature of information. Fuzzy \cite{witte2002fuzzy} and possibilistic belief revision approaches \cite{dubois1993epistemic,dubois1994automated} are also based on the three main principles mentioned before, namely, success, consistency and minimal change. However, fuzzy logic captures imprecision, but not the underlying uncertainty and possibilistic logic is bivalent. Base revision based on an ASP paradigm, that can represent and reason with uncertain and vague information, has not been studied yet.

In this work, we focus on Removed Set Revision (RSR) for base revision with a Unified Answer Set Semantics \cite{paul2019unified}, an interval-valued logic programming paradigm capable of reasoning with fuzzy and uncertain information under a nonmonotonic scenario.

\section{Preliminary Concepts:}

This section focuses on some necessary preliminary concepts, required for further discussion.

\subsection{ASP Base Revision:}

The postulates for characterizing belief base revision over propositional logic have been studied in \cite{hansson2012textbook}. However, logic programs under ASP, being nonmontonic in nature, base revision for ASP is more challenging and requires modified set of postulates. While in propositional case, any subset of a consistent set of sentences is consistent; but for a consistent logic program $Q$, there can be a subset $P \subset Q$ such that $P$ is inconsistent. In other words, the input logic program can be inconsistent but the revision outcome can be consistent. For two answer set programs $P$ and $Q$, let $P*Q$ denotes the revision outcome of $P$ by $Q$, then the necessary set of postulates that characterizes the revision operator "$*$" are as follows \cite{krumpelmann2012belief}:

\textbf{1.Success:}  $Q \subseteq P*Q$

\textbf{2.Inclusion:} $P*Q \subseteq P + Q$

\textbf{3.NM-Consistency:} If there exists some consistent $X$, such that,  $Q \subseteq X \subseteq P \cup Q$ then $P*Q$ is consistent.

\textbf{4.Fullness:} If $r \in (P \cup Q)\setminus (P*Q)$ then $P*Q$ is consistent and $(P*Q) \cup \{r\}$ is inconsistent. 

\textbf{5.Uniformity:} If for all $P' \subseteq P, P' \cup \{Q\}$ is inconsistent if and only if $P' \cup \{R\}$ is inconsistent, then $P \cap (P*Q) = P \cap (P*R)$.

\textbf{6.Weak Disjunction:} If $P = P_1 \cup P_2$ and $P_1$ and $P_2$ have disjoint sets of literals $\mathscr{A}_1$ and $\mathscr{A}_2$ and for each set of literals $\mathscr{A}_r$ of a rule $r \in Q$ it holds $\mathscr{A}_r \cap \mathscr{A}_1 = \phi$ or $\mathscr{A}_r \cap \mathscr{A}_2 = \phi$, then $P*Q \equiv_{\Box} (P_1*Q) \cup (P_2*Q)$.

\textbf{7.Weak Parallelism:} If $Q_1$ and $Q_2$ have disjoint sets of literals $\mathscr{A}_1$ and $\mathscr{A}_2$, and for each set of literals $\mathscr{A}_r$ of a rule $r \in P$ it holds $\mathscr{A}_r \cap \mathscr{A}_1 = \phi$ or $\mathscr{A}_r \cap \mathscr{A}_2 = \phi$, then $P*(Q_1 \cup Q_2) \equiv_{\Box} (P*Q_1)\cup(P*Q_2)$.

where, $P$ and $Q$ are two logic programs and $+$ is a non-closing expansion, such that $P+Q = P \cup Q$. The equivalence relation symbol $\equiv_{\Box}$ is a short-hand representation of a class of equivalence relations between programs. Two special cases of $\equiv_{\Box}$ are the syntactic identity of programs ($\equiv_{P}$) and the equivalence of answer sets of programs ($\equiv_{AS}$). The latter is weaker equivalence than the former.

The aim is to construct a base revision operator satisfying the aforementioned postulates.

\subsection{Removed Set Revision:}

The "removed sets" approach for fusion \cite{hue2008removed} and revision \cite{benferhat2010answer} are proposed for propositional logic and later been extended to ASP \cite{hue2013extending,krumpelmann2012belief}. The basic intuition is that when the added set of formulas is inconsistent with the existing belief base, in order to restore consistency, minimal number of rules from the initial belief base is to be removed. The 'minimality' is determined by some ordering.

\textbf{Definition 1: (Potential Removed Set)} For two logic programs $P$ and $Q$, a set of rules $X$ is a potential removed set if:

(i) $X \subseteq P$

(ii) $(P \setminus X) \cup Q$ is consistent

(iii) for each $X' \subset X$, $(P \setminus X') \cup Q$ is inconsistent.

\textbf{Definition 2 (Preorder and Strategies):}

For two logic programs $P$ and $Q$, let $X$ and $Y$ be potential removed sets for constructing the revised set $P*Q$. Then for every strategy $\mathscr{S}$, a preorder $\leq_{\mathscr{S}}$ over the potential removed sets is defined, such that, $X \leq_{\mathscr{S}} Y$ means $X$ is preferred to $Y$ according to strategy $\mathscr{S}$. 

For base revision strategies $\leq_{\mathscr{S}}$ can be total preorder or partial preorder.

For two logic programs $P$ and $Q$, revising $P$ by $Q$ is providing a new consistent logic program containing $Q$ and differing as little as possible from $P \ \cup \ Q$. In a nonmonotonic scenario following cases may arise:

\textbf{1} $P$ and $Q$ are consistent, but $P \ \cup \ Q$ is inconsistent.

\textbf{2} $P$ and $Q$ are inconsistent, but $P \ \cup \ Q$ is consistent.

When. $P \ \cup \ Q$ is inconsistent, in order to restore consistency, minimal set of rules $X \subseteq P$ is eliminated from $P$ so that $(P\setminus X) \ \cup \ Q$ is consistent. Here, the set $X$ is called the \textbf{Removed Set}. This revision method respects \textbf{consistency}, \textbf{inclusion} and \textbf{principle of minimum information change} of belief base revision.

On a classical setting, the minimality of the Removed Set is measured by set inclusion or by cardinality. In a non-classical framework this notion of minimality is more complex.

\subsection{Unified Answer Set Programs(UnASP):}

In the Unified Answer Set Programming \cite{paul2019unified} framework uncertain vague information can be represented by means of weighted rules and deductive reasoning can be performed with them under nonmonotnonic scenario. In the framework, the set of all sub-intervals of unit interval $[0,1]$ is taken as the set of truth values, i.e., the truth space $\mathscr{T}$. The elements of $\mathscr{T}$ are ordered with respect to the degree of truth and degree of certainty by means of an algebraic structure, namely \textbf{Preorder-based triangle} \cite{ray2018preorder}.

For two elements $[x_1,x_2], [y_1,y_2] \in \mathscr{T}$; the truth-ordering ($\leq_{t_p}$) and knowledge-odering($\leq_{k_p}$) are defined as follows:

$[x_1,x_2] \leq_{t_p} [y_1,y_2]$ iff $\frac{x_1+x_2}{2} \leq \frac{y_1+y_2}{2}$;

$[x_1,x_2] \leq_{k_p} [y_1,y_2]$ iff $(y_2 - y_1) \leq (x_2 - x_1)$.

This algebraic structure is shown to be suitable for performing nonmonotonic reasoning with interval valued truth space; which was not possible with other previously proposed algebraic structures, namely Bilattice-based Triangle \cite{cornelis2007uncertainty}. The logical connectives and negations are defined as follows:

For two elements $[x_1,x_2], [y_1,y_2] \in \mathscr{T}$

1. \textbf{T-norm} $[x_1,x_2] \wedge [y_1,y_2] = [x_1y_1, x_2y_2]$;

2. \textbf{T-conorm} $[x_1,x_2] \vee [y_1,y_2] = [x_1+y_1-x_1y_1, x_2+y_2-x_2y_2]$;

3. \textbf{Classical Negation} $\neg [x_1,x_2] = [1-x_2, 1-x_1]$;

4. \textbf{Negation-as-failure} $not \ [x_1,x_2] = [1-x_1,1-x_1]$.

The syntax of UnASP paradigm consists of finitely many constants and predicate symbols; and infinitely many variables. For an n-ary predicate symbol $p$, $p(t_1,t_2,..,t_n)$ is called an atom, where $t_1,..,t_n$ are variables or constants. If an atom does not contain any variable it is called \textit{grounded}. A \textit{literal} $l$ is either a positive atom or its classical negation. 

A UnASP program is a set of rules of the form:

\begin{center}

$r:a \stackrel{\alpha_r}{\longleftarrow} b_1 \wedge ... \wedge b_k \wedge not \ b_{k+1} \wedge ... \wedge not \ b_n$.

\end{center}

where, $\alpha_r \in \mathscr{T}$ is the weight of the rule, which denotes the epistemic state of the consequent or head of the rule ($a$), when the antecedent or body of the rule ($b_1 \wedge ... \wedge b_k \wedge not \ b_{k+1} \wedge ... \wedge not \ b_n$) is true. The head and the body of rule $'r'$ is denoted by $r(Head)$ and $r(Body)$ respectively. In a rule $b_1,...,b_n$ are positive or negative literals or elements of $\mathscr{T}$ and the rule head $a$ is positive or negated literal. A rule is said to be a \textbf{fact} if $b_i, 1 \leq i \leq n$ are elements of $\mathscr{T}$.

Pieces of information in a knowledge base are not always equally certain. This lack of certainty arises from incomplete evidence, or from conflicting evidence. This notion of certainty is nonprobabilistic and its only purpose is to model the fact that in the knowledge base, some sentences are more disputable or coming from less reliable source due to incomplete information. The rule weight $\alpha_r$ may be used to capture this innate uncertainty levels of various rules. Even, $\alpha_r$ can be used to depict the uncertainty of a rule having exceptions and the degree of uncertainty (length of the interval $\alpha_r$) is meant to summarize these exceptions; e.g., counting them as a surrogate for enumerating them.

 The \textit{atom base} $\textbf{B}_P$ of a program $P$ is the set of all grounded atoms of $P$. $\mathscr{L}_P$ be the set of literals (excluding naf-literals), i.e., $\mathscr{L}_P = \{a|a\in \textbf{B}_p\} \cup \{\neg a|a \in \textbf{B}_p\}$. An \textbf{interpretation}, $I$, is a set $\{l:v_{(I,l)}| l \in \mathscr{L}_P \  \text{and} \  v_{(I,a)} \in \mathscr{T}\}$, which specifies the epistemic states of the literals in the program.

\textbf{Definition 3:} An interpretation $I$ is inconsistent if there exists an atom $a$, such that, $a:v_{(I,a)} \in I$ and $\neg a: v_{(I, \neg a)} \in I$ and $k_{v_{(I,a)}} = k_{v_{(I,\neg a)}}$ but $t_{v_{(I,a)}} \neq 1-t_{v_{(I,\neg a)}}$; where, $k_x(t_x)$ denotes the degree of uncertainty (degree of truth) of some $x \in \mathscr{T}$.

In other words, an inconsistent interpretation assigns contradictory truth status to two complemented literals with same confidence.

The set of interpretations of a program $P$ can be ordered with respect to the uncertainty degree by means of the knowledge ordering ($\leq_{k_p}$). For two interpretations $I$ and $I^*$, $I \leq_{k_p} I^*$ iff $\forall l \in \mathscr{L}_P, v_{(I,l)} \leq_{k_p} v_{(I^*,l)}$. An interpretation $I_k$ is the \textbf{k-minimal} interpretation of a set of interpretations $\Gamma$, iff for no interpretation $I^* \in \Gamma$; $I^* \leq_{k_p} I_k$. If for any $\Gamma$, $I_k$ is unique then it is \textit{k-least}.

\textbf{Definition 4:}
An interpretation $I$ \textit{satisfies} a rule $r$ if for every ground instance of $r$ of the form $r_g:head \stackrel{\alpha_r}{\longleftarrow} body$, $I(head) = (I(body)\wedge \alpha_r)$ or $I(head) >_{k_p} (I(body) \wedge \alpha_r)$ or $I(head) >_{t_P} (I(body) \wedge \alpha_r)$. $I$ is said to be a \textit{model} of a program $P$, if $I$ satisfies every rule of $P$.

\textbf{Definition 5:} A model of a program $P$, $I_m$, is said to be \textbf{supported} iff:

 1. For every grounded rule $r_g: a \stackrel{\alpha_r}{\leftarrow} b$, such that $a$ doesn't occur in the head of any other rule, $I_m(a) = I_m(b)$.

 2. For grounded rules $\{ a \stackrel{\alpha_1}{\leftarrow} b_1, a$     $\stackrel{\alpha_2}{\leftarrow} b_2,..,a \stackrel{\alpha_n}{\leftarrow} b_n\} \ \in P$ having same head $a$, $I_m(a) = (I_m(b_1) \wedge \alpha_1) \vee ... \vee (I_m(b_n) \wedge \alpha_n)$.

 3. For literal $l \in \mathscr{L}$, and grounded rules $r_l: \ l \longleftarrow b_l$, and $\ r_{\neg l}: \neg l \longleftarrow \ b_{\neg l}$, in $P$, $I_m(l) = I_m(b_l) \otimes_K \neg I_m(b_{\neg l})$ and $I_m(b_l) \otimes_k \neg I_m(b_{\neg l})$ exists in $\mathscr{T}$.

The first condition of supportedness guarantees that the inference drawn by a rule is no more certain and no more true than the degree permitted by the rule body and rule weight. The second condition specifies the optimistic way of combining truth assertions for an atom coming from more than one rule. The third condition captures the essence of nonmonotonicity of reasoning. For an atom $a$, rules with $a$ in the head are treated as evidence in favour of $a$ and rules with $\neg a$ in the head stands for evidence against $a$. In such a scenario, the conclusion having more certainty or reliability is taken as the final truth status of $a$.

\textbf{Definition 6:} The \textbf{reduct} of a program $P$ with respect to an interpretation $I$ is defined as:

$P^I = \{r_I:a \stackrel{\alpha_r}{\longleftarrow} b_1 \wedge ... \wedge b_k \wedge not \ I(b_{k+1}) \wedge ... \wedge not \ I(b_n) \ | \ r \in P\}$.

$P^I$ doesn't contain any naf-literal in any rule. For a positive program $P$ (with no rules containing $not$), $P^I = P.$

\textbf{Definition 7:} For any UnASP program $P$, an interpretation $I$ is an \textbf{answer set} if $I$ is an k-minimal supported model of $P^I$. For a positive program the k-minimal model is unique.

The atom not appearing in the head of any rule will be assigned $[0,1]$.

\section{Belief Base revision based on UnAsP}

In this work belief bases are represented using Unified Answer Set Programs (UnASP).

In UnASP the dependencies within a program are complex and cannot be anticipated without considering the input program. The inconsistency of a program $P$ with a new program $Q$ can only be determined by considering $P \cup Q$, as the interaction of rules of both the programs generates inconsistency. Thus, this type of base-revision is \textit{external revision} as the sub-operation takes place outside of the original set.

\subsection{Update of weights of rules with exceptions:}

As mentioned in the previous section, the weight of a rule can be used to signify that it is a disposition \cite{zadeh1985syllogistic}, i.e. a proposition having exceptions and the rule weight summarizes the number of exceptions of a rule by enumerating the exception-capturing rules in the knowledge base. Now if the new knowledge base contains several more exceptions for the same disposition then in the combined program the weight of the disposition has to be updated in order to reflect the modified number of exceptions.

\section{Example 1.}

$P_1 = \{r_{11}: p \stackrel{\alpha}{\leftarrow} q,r$

$r_{12}: r \stackrel{\beta}{\leftarrow} s$

$r_{13}: q \stackrel{\gamma}{\leftarrow}$

$r_{14}: \neg p \longleftarrow t$

$r_{15}: s \stackrel{[1,1]}{\leftarrow} \}$\\

The second program is:

$P_2 = \{r_{21}: p \stackrel{\alpha_1}{\leftarrow} a,b$

$r_{22}: a \stackrel{\beta_1}{\leftarrow} c$

$r_{23}: b \stackrel{\gamma_1}{\leftarrow}$

$r_{24}: \neg p \longleftarrow d\}$

Now, in the programs $P_1$ and $P_2$, rules $r_{11}$ and $r_{21}$ are \textit{dispositions}, with $r_{14}$ and $r_{24}$ pointing their exceptions respectively. The rule weights $\alpha$ and $\alpha_1$ summarize the number of exceptions of rule $r_{11}$ and $r_{21}$ respectively, by enumerating the exception-capturing rules like $r_{14}$ and $r_{24}$. In the program $P_1 \cup P_2$, the weights of rules $r_{11}$ and $r_{21}$ have to be updated in order to enumerate the exceptions combined from both programs, since now in $P_1 \cup P_2$, both the rules $r_{14}$ and $r_{24}$ serve as exceptions for them. The combined program becomes:

$P_1 \cup^* P_2 = \{ r_{1}: p \stackrel{\alpha'}{\leftarrow} q,r$

$r_{2}: p \stackrel{\alpha_1'}{\leftarrow} a,b$

$r_{3}: r \stackrel{\beta}{\leftarrow} s$

$r_{4}: q \stackrel{\gamma}{\leftarrow}$

$r_{5}: s \stackrel{[1,1]}{\leftarrow}$

$r_{6}: a \stackrel{\beta_1}{\leftarrow} c$

$r_{7}: b \stackrel{\gamma_1}{\leftarrow}$

$r_{8}: \neg p \longleftarrow t$

$r_{9}: \neg p \longleftarrow d\}$

Clearly, $\alpha'$ and $\alpha_1'$ are wider intervals than $\alpha$ and $\alpha_1$ respectively, signifying that, with the increase in the number of exceptions in the combined program, the certainties of the dispositions are reduced. 

For two knowledge bases $P$ and $Q$, their union $P \cup^* Q$, with the modified rule weights, is referred to as the \textit{modified union}, to distinguish it from the ordinary union. If no rule weights are modified, then the modified union acts as ordinary union. 

\subsection{Determination of Potential Removed Set:}

After the construction of the modified union of two logic programs, its answer set is to be constructed. The answer set will exist if the modified union is consistent. Otherwise, if the new information is incompatible with the existing knowledge base, no answer set is found. Now in order to restore consistency some rules have to be removed, subject to causal rejection principle \cite{eiter2002properties, osorio2007updates}. The causal rejection principle enforces that in case of conflicts between rules, more recent rules are preferred and older rules are overridden. 

In order to construct the removed set, the following steps are followed.

\subsubsection{Program Transformation:}

\textbf{Definition 8:} For any Unified Answer Set program $P$, the corresponding transformed program $P_T$ is constructed as follows:

(i) For every rule $r \in P$, with weight $\alpha_r$, such that for any other $r' \in P$, $r(Head) \neq r'(Head)$, then the rule $r(Head) \longleftarrow \alpha_r \wedge r(Body) \ $ is included in $P_T$;

(ii) If $r_1,...r_k \in P$, such that $r_1(Head) = r_2(Head)= ... = r_k(Head)$, and weights of $r_1,..,r_k$ are $\alpha_1, ... ,\alpha_k$ respectively, moreover there is no rule $r' \in P$  such that $r'(Head) = \neg r_1(Head)$, then the transformed rule corresponding to $r_1,..,r_k$ is:

$r_T: r_1(Head) \longleftarrow (\alpha_1 \wedge r_1(Body)) \vee ... \vee (\alpha_k \wedge r_k(Body)) \ \in P_T$;

(iii) If program $P$ has rules $r_1,...,r_m$, with weights $\alpha_1,..,\alpha_m$ respectively and $r_1(Head) = ... = r_m(Head)$, and $r_1^{\neg},...,r_n^{\neg}$ with weights $\beta_1,...,\beta_n$ and $r_1^{\neg}(Head) = ... = r_m^{\neg}(Head) = \neg r_1(Head)$; then the transformed rule corresponding to $r_1,..,r_m,r_1^{\neg},..,r_n^{\neg}$ 

$$r_1(Head) \longleftarrow \bigvee_{i=1,..m}(\alpha_i \wedge r_i(Body)) \otimes_k \bigvee_{i=1,..n}(\beta_i \wedge r_i^{\neg}(Body)) \ \in P_T$$

(iv) For each atom $a$ in the Atom base of Program $P$ that does not occur of the rule head of any rule in $P$, a rule $a \longleftarrow [0,1]$ is added to $P_T$.

The operator $\otimes_k$ is a knowledge aggregation operator which takes into account the interaction of epistemic states of an atom and its corresponding negated literal based on their certainty levels. Thus $\otimes_k$ accounts for representing the nonmonotonic relation between an atom and its negation and is defined as follows:

\textbf{Definition 9:} For two intervals $x = [x_1, x_2]$ and $y = [y_1,y_2]$ in $I(L)$; 

\begin{center}

$x \otimes_k y = \begin{cases} x, & \text{if} \ y \leq_{k_p} x \\ y, & \text{if} \ x \leq_{k_p} y \\ [\xi,\xi], & \text{otherwise} \end{cases}$

\end{center}

where $\xi$ is a large positive or negative number and its occurrence denotes that for epistemic states $x, y \in \mathscr{T}$, $x \otimes_k y$ is undefined and hence $x$ and $y$ are contradictory. $\xi$ is chosen to be large so that if it occurs in the body of any rule and undergoes the necessary operations then the head will also be a large number well outside the range of [0,1], signifying the inconsistency.

\textbf{Definition 10 (Transformation Table):} For a UnASP program $P$ and its corresponding transformed program $P_T$, the Transformation Table is a two column table having the rules from $P_T$ in the left column and the associated rules of $P$ from which the transformed rules have been constructed in the right column; i.e. for $r_i \in P_T$ in the $i^{th}$ row of column 1, the $i^{th}$ row of column 2 contains rules from $P$ from which $r_i(Body)$ is constructed.

\textbf{Example 1(continued):}

The program transformation of program $P_1$ is as follows:

$P_1^T = \{r_1^T: p \longleftarrow (\alpha \wedge q \wedge r) \otimes_k \neg t,$

$ \  \  \  \  \  \  \  \  \ r_2^T: r \longleftarrow \beta \wedge s,$

$ \  \  \  \  \  \  \  \  \ r_3^T: q \longleftarrow \gamma,$

$ \  \  \  \  \  \  \  \  \ r_4^T: s \longleftarrow [1,1],$

$ \  \  \  \  \  \  \  \  \ r_5^T: t \longleftarrow [0,1]\}.$

\begin{table*}[h!]
\begin{center}	
\begin{tabular}{|c|c|}

\hline 

Rules from $P^T$ & Rules from $P$\\

\hline

$r_1^T: p \longleftarrow (\alpha \wedge q \wedge r) \otimes_k \neg t$ & $r_{11}: p \stackrel{\alpha}{\leftarrow} q,r; \  r_{14}: \neg p \longleftarrow t$ \\

$r_2^T: r \longleftarrow \beta \wedge s$ & $r_{12}: r \stackrel{\beta}{\leftarrow} s$\\

$r_3^T: q \longleftarrow \gamma$ & $r_{13}: q \stackrel{\gamma}{\leftarrow}$ \\

$r_4^T: s \longleftarrow [1,1]$ & $r_{15}: s \stackrel{[1,1]}{\leftarrow}$\\

$r_5^T: t \longleftarrow [0,1]$ & -\\

\hline

	\end{tabular}
	\end{center}
	\caption{Transformation Table of program $P$}	
	\label{tab:transformation table}
\end{table*}

The transformation table corresponding to the transformation of program $P$ is shown in Table \ref{tab:transformation table}

\subsubsection{Modified Resolution Tree:}

Suppose the program $P$ is to be revised with another program $Q$. During the construction of the answer set of $P \cup^* Q$, programs $P$ and $Q$ conflicts over the epistemic state of an atom $p$, i.e. the model of $P \cup^* Q$ assigns $[\xi,\xi]$ to $p$. Using the transformed program, a modified resolution tree is constructed in order to pinpoint the rules used to derive the epistemic state of the atoms that give rise to inconsistency, i.e. $[\xi, \xi]$, in the models of $P \cup^* Q$. Modified resolution tree is an interval-valued variant of resolution tree for propositional logic. The modified resolution tree showing the derivation of $p$ from program $P$ is constructed by with the following steps:

1. Start with the rule $r_p \in P_T$ having $p$ in the head, i.e., $r_p(Head) = p$.

2. For each atom $a$ in $r_p(Body)$ replace $a$ with the body of the rule $r' \in P_T$, such that $r'(Head) = a$.

3. Step 2 is repeated until every element in $r_p(Body)$ becomes an element of $\mathscr{T}$.

\textbf{Example 1(continued):}

According to the chosen Unified Answer Set semantics the answer set is going to be $\{s:[1,1], b:\gamma_1, q: \gamma, c:[0,1], t:[0,1], r: \beta, a: beta_1 \wedge [0,1], p: ((q\wedge r \wedge \alpha')\vee(a \wedge b \wedge \alpha_1')) \otimes_k (t\vee d)\}$.

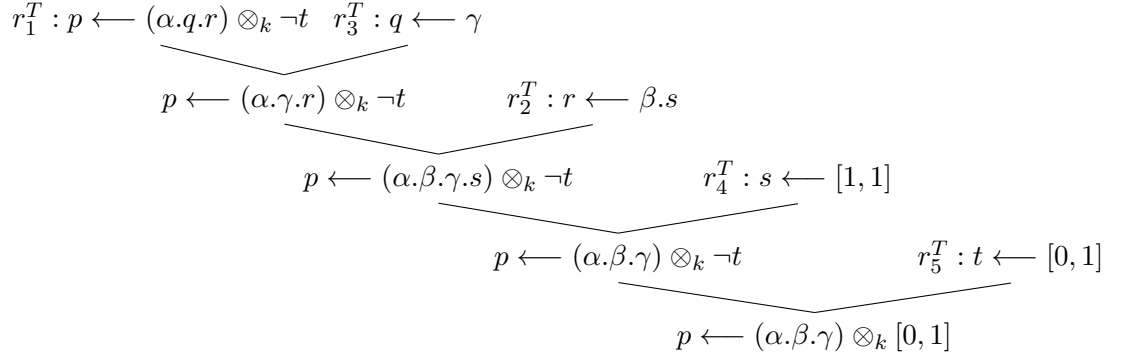
\begin{figure}
\begin{center}
\begin{tikzpicture}[grow'=up]

\Tree [.${p \longleftarrow (\alpha.\beta. \gamma) \otimes_k [0,1]}$ [.${p \longleftarrow (\alpha.\beta. \gamma) \otimes_k \neg t}$ [. {${p \longleftarrow (\alpha . \beta . \gamma . s) \otimes_k \neg t}$} [.${p \longleftarrow (\alpha . \gamma . r) \otimes_k \neg t}$ ${r_1^T:p \longleftarrow (\alpha . q . r) \otimes_k \neg t}$ ${r_3^T: q \longleftarrow  \gamma}$ ]
                           [.${ r_2^T: r \longleftarrow \beta . s }$ ] ]
                           [.{${r_4^T: s \longleftarrow [1,1]}$} ] ] [.{${r_5^T: t \longleftarrow [0,1]}$} ] ]

\end{tikzpicture}
\caption{Resolution Tree}
\label{fig:restree}
\end{center}
\end{figure}

Now suppose, during this revision, programs $P_1$ and $P_2$ assigns contradictory epistemic states to atom p. Hence, in order to find out the removed set, the modified resolution tree for atom $p$ is constructed as shown in Figure \ref{fig:restree}.

\subsubsection{Potential Removed Sets:}

Suppose a belief base $P$, expressed by a UnASP program is revised by another base $Q$. Now, in the combined program $P \cup^* Q$, some of the atoms do not get any epistemic state from $\mathscr{T}$ due to inconsistency.

\textbf{Definition 11:} The \textit{Contradiction Set ($CS_{P\cup^*Q}$)} is the set of all atoms in $\textbf{B}_P \cap \textbf{B}_Q$, that become inconsistent, i.e. get $[\xi,\xi]$. in the answer set of $P \cup^* Q$.

\textbf{Lemma 1:} If $a \in CS_{P\cup^*Q}$ then for any rule $r$ in program $P$ and $Q$, if $a \in r(Body)$ then $r(Head) \in CS_{P\cup^*Q}$.

The proof of the Lemma 1 is straightforward, since if any rule contains an inconsistent atom, having epistemic state $[\xi,\xi]$, then the atom in the head will also become inconsistent as is ensured by the choice of $\xi$.

Now for any atom $p \in CS_{P\cup^*Q}$, the derivation of $p$ gives all the rules of the transformed program $P_T$, that take part in the derivation of $p$. Using the transformation table we can retrieve the actual rules of program $P$ that take part in the derivation and hence construct the \textit{set of potential removed sets} corresponding to $p$, denoted by $PRS_p$. Among all the rules from belief base $P$, that take part in the derivation of $p \in CS_{P\cup^*Q}$, the ones, whose elimination resolves the contradiction, form a potential removed set for $p$ and together they form $PRS_p$, i.e. set of all removed sets. Therefore, essentially $PRS_p$ is a set of rules.

Similarly, for each of the atoms $x \in CS_{P \cup^* Q}$, a set $PRS_x$ is obtained. From the potential removed sets the Removed set is then constructed. 

\textbf{Example 1(continued):} For eliminating the contradiction over the epistemic state of atom $p$, the set of potential removed sets corresponding to $p$, $PRS_p$, is constructed using the modified resolution tree (Figure \ref{fig:restree}) and the transformation Table \ref{tab:transformation table}.

$PRS_p = \{\{r11: p \stackrel{\alpha}{\leftarrow} q,r\}, \{r_{14}: \neg p \longleftarrow t\}, \{r_{12}: r \stackrel{\beta}{\leftarrow} s\}, \{r_{13}: q \stackrel{\gamma}{\leftarrow}\}, \{r_{15}: s \stackrel{[1,1]}{\leftarrow}\}\}$.

\subsection{Strategy for construction of Removed Set:}

\subsubsection{Distance between two set of models in UnASP:}

\textbf{Definition 12:} The \textit{distance} between two elements $[x_1,x_2], [y_1,y_2] \in \mathscr{T}$ is given as:

\begin{center}

$\mathscr{D}([x_1,x_2],[y_1,y_2]) = \frac{|x_1 - x_2| + |y_1 - y_2|}{2}$.

\end{center} 

$\mathscr{D}$ can be used to measure the difference of epistemic states of an atom assigned by two interpretations or models and $0 \leq \mathscr{D}(x,y) \leq 1; \forall x,y \in \mathscr{T}$.

\textbf{Definition 13:} For two interpretations $I$ and $J$, evaluating the set of atoms from some atom base \textbf{B}, $$\mathscr{D}(I,J) = \sum_{\forall a \in \textbf{B}}\mathscr{D}(v_{(I,a)},v_{(J,a)})$$.

For two sets of interpretations $\mathscr{I}$ and $\mathscr{J}$, $$\mathscr{D}(\mathscr{I}, \mathscr{J}) = \text{max}\{\mathscr{D}(I,J)|\forall I \in \mathscr{I}, \forall J \in \mathscr{J}\}$$

\subsubsection{Strategy:}

For two programs $P$ and $Q$, $P \cup^*Q$ is to be constructed. But, if $P \cup^* Q$ gives rise to the contraction set $CS_{P\cup^*Q}$, then the removed set $X$ is constructed based on the following strategies.

1. For $a_1,...,a_k \in CS_{P\cup^*Q}$, with all of $PRS_{a_i} \ (1 \leq i \leq k)$, being mutually disjoint; one rule is to be chosen from each of the $PRS_{a_i}$. For any $a_i$, if all the rules in $PRS_{a_i}$ are totally ordered in terms of their weights with respect to the knowledge ordering ($\leq_{k_p}$), then the $\leq_{k_p}$-least element, i.e., the rule with least certainty, is included in the removed set $X$.

2. Say, for any two $a_i, a_j \in CS_{P\cup^*Q}$, the sets $PRS_{a_i}$ and $PRS_{a_j}$ overlap, and the set of rules $PRS_{a_i} \cap PRS_{a_j}$ is totally ordered with respect to the knowledge ordering ($\leq_{k_p}$) of the weights of the rules, then the least element in the order is included in $X$. This single rule eliminates the contradiction for both $a_i$ and $a_j$ in the model of $P \cup^* Q$.

3. If in the above two cases, the rules form a partial preorder, with more than one minimal elements (with respect to $\leq_{k_p}$), then more than one removed sets can be obtained; each of which respects the principle of minimal change in a syntactic way, i.e. contains minimum number of rules required to restore consistency of $P \cup^* Q$. To choose one from these syntactically minimal removed sets a distance-based criteria is imposed on the models to ensure minimality in a semantic way. Among all the syntactically minimal removed sets a particular set $X$ is chosen to be the removed set if $\mathscr{D}(As_P, As_{(P \setminus X)})$ is minimum, i.e., the answer sets of $P \setminus X$ is "closest" to the answer sets of $P$.

\subsection {Belief Base Revision Operator:}

Depending on the base revision strategy, described in the previous subsection, a base revision operator ($*_u$) is defined for knowledge bases represented with UnASP logic programs. 

\textbf{Definition 14:} Let P,Q be two logic programs. Let $\mathscr{X}_{P,Q}$ be the set of removed sets and $f$ be a selection function which chooses a particular removed set $X_{P,Q}$ from $\mathscr{X}_{P,Q}$, i.e., $f(\mathscr{X}_{P,Q}) = X_{P,Q}$. The revision operator $*_u$ is a function from $P\times P$ to $P$, such that $P*_uQ = (P\setminus f(\mathscr{X}_{P,Q})) \cup Q$.

\textbf{Example 1(continued):} In the example consider a specific case where, $\alpha \leq_{k_p} \gamma \leq_{k_p} \beta \leq_{k_p} [1,1]$, (i.e., a total order), then $\mathscr{X}_{P_1,P_2}$ is singleton and is $\{p \stackrel{\alpha}{\leftarrow} q,r\}$, i.e. the least certain rule is eliminated. Also $X_{P_1,P_2} = f(\mathscr{X}_{P_1,P_2}) = \{p \stackrel{\alpha}{\leftarrow} q,r\}$.

Hence, $P_1*_uP_2 = (P_1 \setminus \{p \stackrel{\alpha}{\leftarrow} q,r\}) \cup P_2$.

\section{Characterization of the base revision operator with respect to the revision postulates:}

This section investigates whether the base revision operator developed in the previous section satisfies the necessary postulates mentioned in Section 2.

\textbf{1. Success:}

\textbf{Proposition:} If program $P$ doesn't contain the exceptions of any dispositiond of program $Q$, then $Q \subseteq P \cup^* Q$ and also $Q \subseteq P *_u Q$ (since the removed set doesn't contain any rule from $Q$). Hence Success postulate is respected by the base revision operator $*_u$.

However, when $P$ contains exceptions of some of the rules of $Q$, then while combining two knowledge bases $P$ and $Q$, due to the interaction of dispositions and exceptions, weights of dispositions in $P$ and $Q$ are modified. Therefore, some rules of $Q$ being permanently modified, and Success postulate is not strictly satisfied. But the essence of the postulate is preserved. All the rules present in $Q$ are also present in $P *_u Q$; but the rule weights may alter.

\textbf{2. Inclusion:}

The form of Inclusion postulate that is satisfied is as follows:

\begin{center}

$P*_uQ \subseteq P \cup^* Q$

\end{center} 

where, $\cup^*$ differs from ordinary union, denoted by $+$ in Section 2, in terms of rule weights only.

\textbf{3. NM Consistency:}

This postulate encompasses both of the cases when the new knowledge base $Q$ is consistent or not. If $Q$ is consistent then the consistency of $P*_uQ$ depends on the removal strategy.

If $Q$ is inconsistent then the following can be stated.

\textbf{Theorem 1:} If there exists some consistent $X$, such that, $Q \subseteq X \subseteq P\cup^*Q$, and program $P$ has some rules $r_1,..,r_i$ so that $CS_Q \subseteq \{r_1(Head),..,r_i(Head)\}$, then $P*_uQ$ can be consistent.

\textbf{Proof:} If $Q$ is inconsistent then $CS_Q$ contains the atoms which gets $[\xi,\xi]$ in all the answer sets of $Q$. Since, the success postulate ensures that no rule is removed from $Q$ in $P*_uQ$, the inconsistency of $Q$ persists in $P*_uQ$ unless $P$ contains some rules having atoms from $CS_Q$ in their heads. In such a scenario following two events can take place:

(1)If $P$ contains some exceptions for the dispositions whose heads are atoms from $CS_Q$, then weights of rules $r \in Q$ with $r(Head) \in CS_Q \cap \textbf{B}_P$ are updated and the contradiction of $Q$ is eliminated.

(2)If no rule weight of $Q$ is updated in $P\cup^*Q$, then also the rules in $P$ alters the contradictory epistemic state of atoms in $CS_Q$. Suppose for $a \in CS_{Q}$ and in some supported interpretation (Definition 5) the epistemic state of $a$ becomes $a:x \otimes_k \neg x$ for some $x \in \mathscr{T}$. If $P$ contains a rule $a \longleftarrow y$ with $y \in \mathscr{T}$ then in the supported model of $P\cup^*Q$ the epistemic state of $a$ becomes $(x \vee y) \otimes_k \neg x$. The disjunction $\vee$ being dual of the product conjunction, $x \vee y \neq x$, unless $ x = [1,1]$. Thus the contradiction is removed unless $x = [1,1]$. However, if $x = [1,1]$, i.e, for any atom $a$ the contradiction is of the form $a: [1,1] \otimes_k [0,0]$ then $P$ cannot resolve this contradiction and a consistent $P *_u Q$ can not be constructed. (\textbf{Q.E.D})

\textbf{4. Fullness:}

 \textbf{Theorem 2:} For any rule $r \in (P \cup^* Q)\setminus(P *_u Q)$, then $P *_u Q$ is consistent and $(P *_u Q) \cup \{r\}$ is inconsistent.

\textbf{Proof:} Any rule $r \in (P \cup^* Q)\setminus(P *_u Q)$ comes from $PRS_a$ for some atom $a \in CS_{P\cup^*Q}$. Therefore, $(P *_u Q) \cup \{r\}$ contradicts over the epistemic state of $a$ and hence is inconsistent. (\textbf{Q.E.D})

\textbf{5. Uniformity:}

\textbf{Theorem 3}: If for all $P' \subseteq P$, $P' \cup^* \{Q\}$ is inconsistent iff $P' \cup^* \{R\}$ is inconsistent, then $P \cap (P*_u Q) = P \cap (P*_uR)$; i.e., for revising with respect to $Q$ and $R$ same set of rules from $P$ is retained.

\textbf{Proof:} It is given that for any $P' \subseteq P$, $P' \cup^* \{Q\}$ is inconsistent iff $P' \cup^* \{R\}$ is inconsistent.

\textbf{Claim 1:} While revising $P$ with $Q$ and $P$ with $R$, $P\cup^*Q$ and $P\cup^*R$ contradicts over same set of atoms, i.e., $CS_{P\cup^*Q} = CS_{P\cup^*R}$.

Following Claim 1, since $CS_{P\cup^*Q} = CS_{P\cup^*R}$, both generates the same $PRS$ and accordingly the removed sets, which are solely dependent on $P$ (and not on $Q$ or $R$), will be same as well for both the cases. Hence $P \cap (P*_u Q) = P \cap (P*_uR) = P \setminus X$ (where, $X$ is the removed set form $P$).

\textbf{Proof of Claim 1:} Suppose not. Assume an atom $a \in CS_{P \cup^* Q}$ and $a \notin CS_{P \cup^* R}$. Construct $P' \subseteq P$, with all rules (i) having $a$ in their heads (ii) All the rules associated in the modified resolution tree of $a$, i.e. all rules used in the derivation of the epistemic state of of $a$. Now clearly $P' \cup^* \{Q\}$ is inconsistent (w.r.t $a$) but $P' \cup^* \{R\}$ is not; because if $P' \cup^* \{R\}$ were inconsistent for some other atom, say $b$, then $b$ would appear in the resolution tree of $a$ and eventually $a \in CS_{P*_uR}$ (from Lemma 1). So, this contradicts the assumption. Hence Claim 1 is proved. (\textbf{Q.E.D})

\textbf{6. Weak Disjunction:} 

The disjunction principle is too strong for base revision with logic programs and hence is weakened \cite{krumpelmann2012belief}. For the base revision strategy, developed here, weak disjunction holds if in the new program $Q$ there is no disposition whose exceptions belong to the original program $P$, i.e., no weight update is required for $Q$.

\textbf{Theorem 4:} If $P = P_1 \cup P_2$ and $P_1$ and $P_2$ have disjoint sets of literals $\mathscr{A}_1$ and $\mathscr{A}_2$ and for each set of literals $\mathscr{A}_r$ of a rule $r \in Q$ it holds $\mathscr{A}_r \cap \mathscr{A}_1 = \phi$ or $\mathscr{A}_r \cap \mathscr{A}_2 = \phi$, and $Q$ \textit{does not contain any disposition whose exceptions belong to $P$}, then $P*_uQ \equiv_{P} (P_1*_uQ) \cup (P_2*_uQ)$. 

\textbf{Proof:} $Q$ is partitioned into disjoint sub-programs $Q_1$ and $Q_2$; i.e. $Q = Q_1 \cup Q_2$, so that $Q_1 = \{r \in Q | \mathscr{A}_r \cap \mathscr{A}_2 = \phi\}$ and $Q_2 = \{r \in Q | \mathscr{A}_r \cap \mathscr{A}_1 = \phi\}$. So, the sub-program $Q_1$ does not interact with $P_2$ and $Q_2$ does not interact with $P_1$. 

Suppose, $CS_{P*_uQ} = \{a_1,...,a_n\}$, and $a_1,..,a_i \in \mathscr{A}_1$ and $a_j,...,a_n \in \mathscr{A}_2$. $P_1$ and $P_2$ being disjoint, the removed set $X_1$, corresponding to $a_1,..,a_i$ comes from $P_1$ and similarly $X_2$ comes from $P_2$.

So, $P*_uQ \equiv_{P} (\{P_1 \setminus X_1\} \cup \{P_2\setminus X_2\}) \cup \{Q\}$

$ \  \  \  \  \  \  \  \  \  \  \ \equiv_{P} (\{P_1 \setminus X_1\} \cup \{Q\}) \cup (\{P_2\setminus X_2\} \cup \{Q\})$

$ \  \  \  \  \  \  \  \  \  \  \ \equiv_{P} (P_1 *_u Q) \cup (P_2 *_u Q)$. (\textbf{Q.E.D})

Therefore, though the condition for weak disjunction is made more strict but the equivalence is syntactical program equivalence, i.e. the strongest.

However, if $P$ contains some of the exceptions of some dispositions of $Q$, then the corresopnding rule weights are updated. In that case we obtain $X_{P,Q} = X_{P_1,Q} \cup X_{P_2,Q}$, i.e.

\begin{center}

$P \setminus (P*_u Q) = (P_1 \setminus (P_1*_u Q)) \cup (P_2 \setminus (P_2 *_u Q))$.

\end{center}

\textbf{7. Weak Parallelism:} 

Weak Parallelism can be expressed in terms of removed sets.

\textbf{Theorem 5:} If $Q = Q_1 \cup Q_2$ so that the set of literals $\mathscr{A}_1$ and $\mathscr{A}_2$ respectively and $\mathscr{A}_1 \cap \mathscr{A}_2 = \phi$. For each set of literals $\mathscr{A}_r$ of a rule $r \in P$ if $\mathscr{A}_r \cap \mathscr{A}_1 = \phi$ or  $\mathscr{A}_r \cap \mathscr{A}_2 = \phi$; then then $P*_u(Q_1 \cup Q_2) \equiv_{P} (P*_uQ_1)\cup(P*_uQ_2)$.

\textbf{Proof:} The program $P$ can be partitioned into two disjoint sub-programs, as $P = P_1 \cup P_2$, so that $P_1 = \{r \in P|\mathscr{A}_r \cap \mathscr{A}_2 = \phi\}$ and $P_2 = \{r \in P|\mathscr{A}_r \cap \mathscr{A}_1 = \phi\}$. Now $P_i$ and $Q_j$ are disjoint if $i \neq j$. Moreover, $Q_1, Q_2$ being disjoint, $CS_{P\cup^*Q_1}$ and $CS_{P\cup^*Q_2}$ are disjoint and we have $CS_{P\cup^*Q_1} = CS_{P_1\cup^*Q_1}$ and $CS_{P\cup^*Q_2} = CS_{P_2\cup^*Q_2}$. So $\mathscr{X}_{P,Q_1} = \mathscr{X}_{P_1,Q_1}$ and $\mathscr{X}_{P,Q_2} = \mathscr{X}_{P_2,Q_2}$. If it is assumed that the selection function $f$ selects same removed sets then we have, $X_{P,Q_1} = X_{P_1,Q_1} \subseteq P_1$ and $X_{P,Q_2} = X_{P_2,Q_2} \subseteq P_2$.

$P*_u(Q_1 \cup Q_2) = \{P\setminus X_{P,(Q_1 \cup Q_2)}\} \cup {Q_1 \cup Q_2}$

$ \  \  \  \  \  \  \  \  \  \  \  \  \  \  \  \ = P\setminus X_{P_1,Q_1}\cup X_{P_2,Q_2} \cup Q_1 \cup Q_2$  

$ \  \  \  \  \  \  \  \  \  \  \  \  \  \  \  \  \ = (P_1 \setminus X_{P_1,Q_1}) \cup (P_2 \setminus X_{P_2,Q_2}) \cup Q_1 \cup Q_2$

$ \  \  \  \  \  \  \  \  \  \  \  \  \  \  \  \  \ = ((P_1 \setminus X_{P_1,Q_1}) \cup Q_1) \cup ((P_2 \setminus X_{P_2,Q_2})  \cup Q_2)$

$ \  \  \  \  \  \  \  \  \  \  \  \  \  \  \  \  \ = (P_1 *_u Q_1) \cup (P_2 *_u Q_2)$

$ \  \  \  \  \  \  \  \  \  \  \  \  \  \  \  \  \ = (P *_u Q_1) \cup (P *_u Q_2)$

$P*_u(Q_1 \cup Q_2) \equiv_P (P *_u Q_1) \cup (P *_u Q_2)$. \textbf{Q.E.D}

So, Inclusion, Fullness, Uniformity and Weak Parallelism postulates are satisfied by the base revision operator $*_u$. The rest of the postulates are satisfied with some minor modifications, though the modifications don't alter the essence of those postulates.

\section{Conclusion:}

Belief revision is an indispensable aspect of nonmonotonic reasoning. When an intelligent agent obtains new knowledge consistent with its current information that knowledge is added to the agent's knowledge base. If however the new knowledge contradicts the agent's current information, a method for incorporating this new knowledge has to be developed. In this work the knowledge base is represented using a UnASP program and a base revision operator based on removed set revision strategy is developed. The revision operator exploits the knowledge ordering of the underlying preorder-based triangle for achieving the minimality of removed set. The operation, developed here, is very intuitive and its performance is satisfactory as it respects the essence of all the required postulates for base revision. 

\textbf{Acknowledgments:} The second author acknowledges the research grant received from Department of Science and Technology, Government of India, in the form of INSPIRE Fellowship.

\bibliographystyle{spmpsci}      % basic style, author-year citations
\bibliography{biblist2}

\end{document}